\documentclass[10pt,twocolumn,letterpaper]{article}

\usepackage{cvpr}
\usepackage{epsfig}
\usepackage{graphicx}
\usepackage{subfigure}
\usepackage{amsmath}
\usepackage{amssymb}
\usepackage{amsthm}
\usepackage{amsfonts}
\usepackage{mathrsfs}
\usepackage{booktabs}
\usepackage{multirow, eucal}

\usepackage{helvet}
\usepackage{courier}
\usepackage{bm}
\usepackage{graphicx}
\usepackage{color}
\usepackage{epstopdf}
\usepackage{wrapfig}
\usepackage{picinpar}
\usepackage{url}

\usepackage[vlined,ruled,linesnumbered]{algorithm2e}
\usepackage{cite}

\usepackage{caption}
\cvprfinalcopy

\def\etal{{\em et al.\/}\, }

\def\0{{\bf 0}}
\def\1{{\bf 1}}

\def\citep{\cite}
\def\citet{\cite}

\def\etal{{\em et al.\/}\, }

\begin{document}

\title{Towards Effective Low-bitwidth Convolutional Neural Networks\thanks{B. Zhuang, C. Shen, L. Liu and I. Reid are with The University of Adelaide,
  Australia. M. Tan is with South China University of Technology, China.}
  \thanks{Correspondence to C. Shen (e-mail: chhshen@gmail.com).}
}

\author{
Bohan Zhuang, Chunhua Shen,
Mingkui Tan, Lingqiao Liu,  Ian Reid
}

\maketitle

\begin{abstract}
	This paper tackles the problem of training a deep convolutional neural network with both low-precision weights and low-bitwidth activations. Optimizing a low-precision network is very challenging since the training process can easily get trapped in a poor local minima, which results in substantial accuracy loss. To mitigate this problem, we propose three simple-yet-effective approaches to improve the network training.
	First, we propose to use a two-stage optimization strategy to progressively find good local minima. Specifically, we propose to first optimize a net with quantized weights and then quantized activations. This is in contrast to the traditional methods which optimize them simultaneously.
	Second, following a similar spirit of the first method, we propose another progressive optimization approach which progressively decreases the bit-width from high-precision to low-precision during the course of training.
	Third, we adopt a novel learning scheme to jointly train a full-precision model alongside the low-precision one. By doing so, the full-precision model provides hints to guide the low-precision model training.
	Extensive experiments on various datasets (\ie, CIFAR-100 and ImageNet) show the effectiveness of the proposed methods. To highlight, using our methods to train a 4-bit precision network leads to no performance decrease in comparison with its full-precision counterpart with standard network architectures (\ie, AlexNet and ResNet-50).

\end{abstract}

\tableofcontents
\clearpage

	\section{Introduction}
The state-of-the-art deep neural networks ~\cite{krizhevsky2012imagenet, simonyan2014very, he2016deep} usually involve millions of parameters and need billions of FLOPs during computation. Those memory and computational cost can be unaffordable for mobile hardware device or especially implementing deep neural networks on chips. To improve the computational and memory efficiency, various solutions have been proposed, including pruning network weights \cite{han2015learning, han2015deep}, low rank approximation of weights \cite{kim2015compression, zhang2016accelerating}, and training a low-bit-precision network \cite{zhou2017incremental, courbariaux2015binaryconnect, zhu2016trained, zhou2016dorefa}. In this work, we follow the idea of training a low-precision network and our focus is to improve the training process of such a network. Note that in the literature, many works adopt this idea but only attempt to quantize the weights of a network while keeping the activations to 32-bit floating point~\cite{zhou2017incremental, courbariaux2015binaryconnect, zhu2016trained}. Although this treatment leads to lower performance decrease comparing to its full-precision counterpart, it still needs substantial amount of computational resource requirement to handle the full-precision activations. Thus, our work targets the problem of training network with \textit{both low-bit quantized weights and activations.}

The solutions proposed in this paper contain three components. They can be applied independently or jointly. The first method is to adopt a two-stage training process. At the first stage, only the weights of a network is quantized. After obtaining a sufficiently good solution of the first stage, the activation of the network is further required to be in low-precision and the network will be trained again. Essentially, this progressive approach first solves a related sub-problem, i.e., training a network with only low-bit weights and the solution of the sub-problem provides a good initial point for training our target problem. Following the similar idea, we propose our second method by performing progressive training on the bit-width aspect of the network. Specifically, we incrementally train a serial of networks with the quantization bit-width (precision) gradually decreased from full-precision to the target precision. The third method is inspired by the recent progress of mutual learning \cite{zhang2017deep} and information distillation \cite{romero2014fitnets, hinton2015distilling, parisotto2016actor, zagoruyko2016paying, ba2014deep}. The basic idea of those works is to train a target network alongside another guidance network. For example, The works in \cite{romero2014fitnets, hinton2015distilling, parisotto2016actor, zagoruyko2016paying, ba2014deep} propose to train a small student network to mimic the deeper or wider teacher network. They add an additional regularizer by minimizing the difference between student's and teacher's posterior probabilities~\cite{hinton2015distilling} or intermediate feature representations~\cite{ba2014deep, romero2014fitnets}. It is observed that by using the guidance of the teacher model, better performance can be obtained with the student model than directly training the student model on the target problem.  Motivated by these observations, we propose to train a full-precision network alongside the target low-precision network. Also, in contrast to standard knowledge distillation methods, we do not require to pre-train the guidance model. Rather, we allow the two models to be trained jointly from scratch since we discover that this treatment enables the two nets adjust better to each other. 

Compared to several existing works that achieve good performance when quantizing both weights and activations~\cite{wu2016quantized, zhou2016dorefa, hubara2016binarized, rastegari2016xnor}, our methods is more considerably scalable to the deeper neural networks \cite{he2016deep, he2016identity}. For example, some methods adopt a layer-wise training procedure \cite{wu2016quantized}, thus their training cost will be significantly increased if the number of layers becomes larger. In contrast, the proposed method does not have this issue and we have experimentally demonstrated that our method is effective with various depth of networks (\ie, AlexNet, ResNet-50).

	\section{Related work}
\label{sec:related_work}

Several methods have been proposed to compress deep models and accelerate inference during testing. We can roughly summarize them into four main categories: limited numerial percision, low-rank approximation, efficient architecture design and network pruning. 

\textbf{Limited numerical precision}
When deploying DNNs into hardware chips like FPGA, network quantization is a must process for efficient computing and storage. Several works have been proposed to quantize only parameters with high accuracy~\cite{courbariaux2015binaryconnect, zhu2016trained,  zhou2017incremental}. Courbariaux \etal \cite{courbariaux2015binaryconnect} propose to constrain the weights to binary values (\ie, -1 or 1) to replace multiply-accumulate operations by simple accumulations. To keep a balance between the efficiency and the accuracy, ternary networks~\cite{zhu2016trained} are proposed to keep the weights to 2-bit while maintaining high accuracy. Zhou \etal \cite{zhou2017incremental} present incremental network quantization (INQ) to efficiently convert any pre-trained full-precision CNN model into low-precision whose weights are constrained to be either powers of two or zero. Different from these methods, a mutual knowledge transfer strategy is proposed to jointly optimize the full-precision model and its low-precision counterpart for high accuracy. What's more, we propose to use a progressive optimization approach to quantize both weights and activations for better performance. 

\textbf{Low-rank approximation}
Among existing works, some methods attempt to approximate low-rank filters in pre-trained networks~\cite{kim2015compression, zhang2016accelerating}. In~\cite{zhang2016accelerating}, reconstruction error of the nonlinear responses are minimized layer-wisely, with subject to the low-rank constraint to reduce the computational cost. 
Other seminal works attempt to restrict filters with low-rank constraints during training phrase~\cite{novikov2015tensorizing, tai2015convolutional}. To better exploit the structure in kernels, it is also proposed to use low-rank tensor decomposition approaches~\cite{denton2014exploiting, novikov2015tensorizing} to remove the redundancy in convolutional kernels in pretrained networks.

\textbf{Efficient architecture design}
The increasing demand for running highly energy efficient neural networks for hardware devices have motivated the network architecture design. GoogLeNet~\cite{szegedy2015going} and SqueezeNet~\cite{iandola2016squeezenet} propose to replace 3x3 convolutional filters with 1x1 size, which tremendously increase the depth of the network while decreasing the complexity a lot. ResNet~\cite{he2016deep} and its variants~\cite{zagoruyko2016wide, he2016identity} utilize residual connections to relieve the gradient vanishing problem when training very deep networks. Recently, depthwise separable convolution employed in Xception~\cite{chollet2016xception} and  MobileNet~\cite{howard2017mobilenets} have been proved to be quite effective. Based on it, ShuffleNet~\cite{zhang2017shufflenet} generalizes the group convolution and the depthwise separable convolution to get the state-of-the-art results. 

\textbf{Pruning and sparsity}
Substantial effort have been made to reduce the storage of deep neural networks in order to save the bandwidth for dedicated hardware design. Han~\etal \cite{han2015learning, han2015deep} introduce ``deep compression'', a three stage pipeline: pruning, trained quantization and Huffman coding to effectively reduce the memory requirement of CNNs with no loss of accuracy. Guo~\etal \cite{guo2016dynamic} further incorporate connection slicing to avoid incorrect pruning. More works~\cite{wen2016learning, lebedev2016fast, liu2015sparse} propose to employ structural sparsity for more energy-efficient compression.

\section{Methods}
\begin{figure*}[!t]
	\centering
	\resizebox{0.9\linewidth}{!}
	{
		\begin{tabular}{c}
			\includegraphics{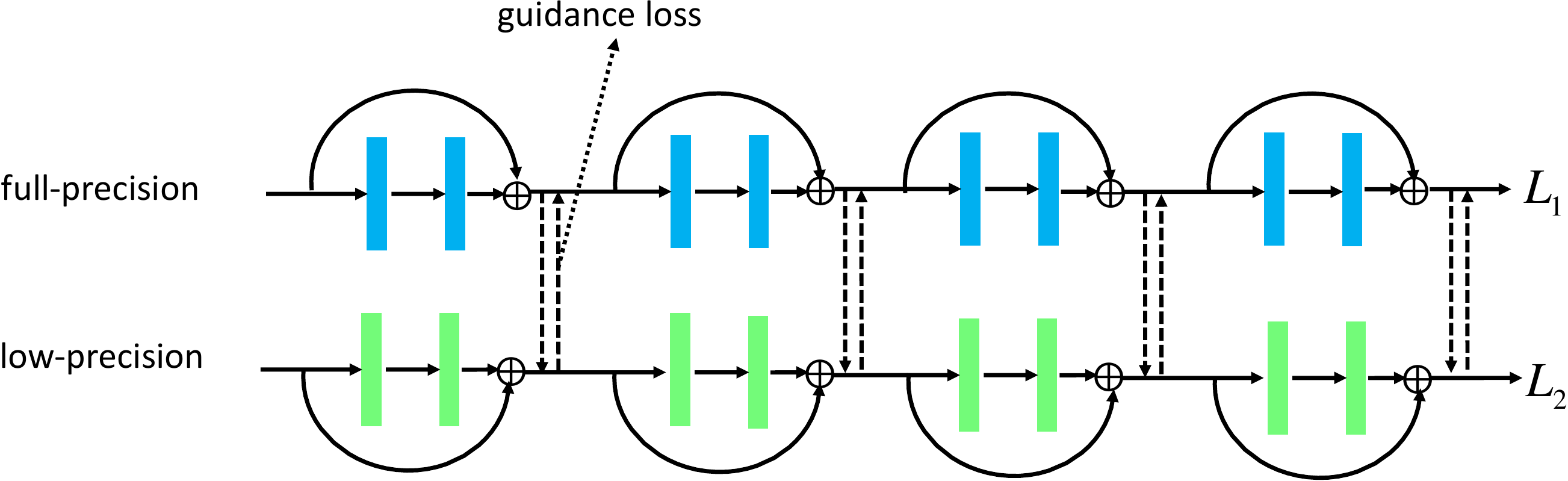}
		\end{tabular}
	}
	\caption{Demonstration of the guided training strategy. We use the residual network structure for illustration.}
	\label{fig:knowledge_transfer}
\end{figure*}
In this section, we will first revisit the quantization function in the neural network and the way to train it. Then we will elaborate our three methods in the subsequent sections.
\subsection{Quantization function revisited} \label{sec:baseline}

A common practise in training a neural network with low-precision weights and activations is to introduce a quantization function. Considering the general case of $k$-bit quantization as in~\cite{zhou2016dorefa}, we define the quantization function $Q(\cdot)$ to be
\begin{equation}
	{z_q} = Q({z_r}) = \frac{1}{{{2^k} - 1}}round(({2^k} - 1){z_r})
\end{equation}
where ${z_r} \in [0,1]$ denotes the full-precision value and ${z_q} \in [0,1]$ denotes the quantized value. With this quantization function, we can define the weight quantization process and the activation quantization process as follows:

\noindent \textbf{Quantization on weights}:
\begin{equation}\label{eq:quan-weigtht}
	{w_q} = Q(\frac{{\tanh (w)}}{{2\max (\left| {\tanh (w)} \right|)}} + \frac{1}{2}).
\end{equation}In other words, we first use $\frac{{\tanh (w)}}{{2\max (\left| {\tanh (w)} \right|)}} + \frac{1}{2}$ to obtain a normalized version of $w$ and then perform the quantization, where $\tanh(\cdot)$ is adopted to reduce the impact of large values.

\noindent \textbf{Quantization on activations}:

 Same as \cite{zhou2016dorefa}, we first use a clip function $f(x) = clip(x,\,0,1)$ to bound the activations to $[0, 1]$. After that, we conduct quantize the activation by applying the quantization function $Q(\cdot)$ on $f(x)$.
\begin{equation} \label{eq:quan-activations}
	{x_q} = Q(f(x)).
\end{equation}

\noindent \textbf{Back-propagation with quantization function}: In general, the quantization function is non-differentiable and thus it is impossible to directly apply the back-propagation to train the network. To overcome this issue, we adopt the straight-through estimator \cite{zhou2016dorefa, hubara2016binarized, bengio2013estimating} to approximate the gradients calculation. Formally, we approximate the partial gradient $\frac{{\partial {z_q}}}{{\partial {z_r}}}$ with an identity mapping, namely $\frac{{\partial {z_q}}}{{\partial {z_r}}} \approx 1$.  Accordingly, $\frac{{\partial l}}{{\partial {z_r}}}$ can be approximated by
\begin{equation}
	\frac{{\partial l}}{{\partial {z_r}}} = \frac{{\partial l}}{{\partial {z_q}}}\frac{{\partial {z_q}}}{{\partial {z_r}}} \approx \frac{{\partial l}}{{\partial {z_q}}}.
\end{equation}

\subsection{Two-stage optimization}\label{sec:two-stage}
With the straight-through estimator, it is possible to directly optimize the low-precision network. However, the gradient approximation of the quantization function inevitably introduces noisy signal for updating network parameters. Strictly speaking, the approximated gradient may not be the right updating direction. Thus, the training process will be more likely to get trapped at a poor local minima than training a full precision model. Applying the quantization function to both weights and activations further worsens the situation.

To reduce the difficulty of training, we devise a two-stage optimization procedure: at the first stage, we only quanitze the weights of the network while setting the activations to be full precision. After the converge (or after certain number of iterations) of this model, we further apply the quantization function on the activations as well and retrain the network. Essentially, the first stage of this method is a related subproblem of the target one. Compared to the target problem, it is easier to optimize since it only introduces quantization function on weights. Thus, we are more likely to arrive at a good solution for this sub-problem. Then, using it to initialize the target problem may help the network avoid poor local minima which will be encountered if we train the network from scratch.
Let $M_{low}^{K}$ be the high-precision model with $K$-bit. We propose to learn a low-precision model $M_{low}^{k}$ in a two-stage manner with $M_{low}^{K}$ serving as the initial point, where $k<K$.
 The detailed algorithm is shown in Algorithm \ref{algo:two-stage}.
\begin{algorithm}[]
	\KwIn{Training data $\{ ({{\bf{x}}_i},y_i)\}_{i=1}^N$; A $K$-bit precision model $M_{low}^K$.}
	\KwOut{A low-precision deep model $M^k_{low}$ with weights ${{\bf{W}}_{low}}$ and activations being quantized into $k$-bit.}

	\textbf{Stage 1}: Quantize ${{\bf{W}}_{low}}$:\\
	\For{ $\mathrm{epoch} = 1,...,L$}
	{
		\For{ $t = 1,...T$}
		{
			Randomly sample a mini-batch data;\\
			Quantize the weights ${{\bf{W}}_{low}}$ into $k$-bit by calling some quantization methods with $K$-bit activations\;
		}
	}
	\textbf{Stage 2}: Quantize activations:\\
	Initialize ${{\bf{W}}_{low}}$ using the converged $k$-bit weights from \textbf{Stage 1} as the starting point; \\
	\For{ $\mathrm{epoch} = 1,...,L$}
	{
		\For{ $t = 1,...T$}
		{
			Randomly sample a mini-batch data;\\
			Quantize the activations into $k$-bit  by calling some quantization methods while keeping the weights to $k$-bit;
		}
	}
	\caption{Two-stage optimization for $k$-bit quantization}
	\label{algo:two-stage}
\end{algorithm}

\subsection{Progressive quantization} \label{sec:progressive}

The aforementioned two-stage optimization approach suggests the benefits of using a related easy optimized problem to find a good initialization. However, separating the quantization of weights and activations is not the only solution to implement the above idea. In this paper, we also propose another solution which progressively lower the bitwidth of the quantization during the course of network training.
Specifically, we progressively conduct the quantization from higher precisions to lower precisions (\eg, 32-bit $\to$ 16-bit $\to$ 4-bit $\to$ 2-bit). The model of higher precision will be used the the starting point of the relatively lower precision, in analogy with annealing.

Let $\{{b_1},...,{b_n}\}$ be a  sequence precisions, where  $b_n<b_{n-1}, ..., b_2<{b_1}$, $b_n$ is the target precision and $b_1$ is set to 32 by default. The whole progressive optimization procedure  is summarized in as Algorithm~\ref{algo:progressive optimization}.
 Let $M_{low}^{k}$ be the low-precision model with $k$-bit and $M_{full}$ be the full precision model. In each step, we propose to learn $M_{low}^{k}$, with the solution in the $(i-1)$-th step, denoted by $M_{low}^{K}$, serving as the initial point, where $k<K$.
\begin{algorithm}[]
	\KwIn{Training data $\{ ({{\bf{x}}_j},y_j)\}_{j=1}^N$; A pre-trained 32-bit full-precision  model ${M_{full}}$ as baseline; the precision sequence $\{{b_1},...,{b_n}\}$ where $b_n<b_{n-1}, ..., b_2<{b_1} = 32$.}
	\KwOut{A low-precision deep model $M_{low}^{b_n}$.}
	Let $M_{low}^{b_1}=M_{full}$, where $b_1 = 32$\;
	\For{ $i = 2,...n$}
	{
		Let $k = b_i$ and $K=b_{i-1}$\;
		Obtain $M_{low}^{k}$ by calling some quantization methods with $M_{low}^{K}$  being the input\;
	}
	\caption{Progressive quantization for accurate CNNs with low-precision weights and activations}
	\label{algo:progressive optimization}
\end{algorithm}

\subsection{Guided training with a full-precision network}\label{sec:mutual}
The third method proposed in this paper is inspired by the success of using information distillation ~\cite{romero2014fitnets, hinton2015distilling, parisotto2016actor, zagoruyko2016paying, ba2014deep} to train a relatively shallow network. Specifically, these methods usually use a teacher model (usually a pretrained deeper network) to provide guided signal for the shallower network. Following this spirit, we propose to train the low-precision network alongside another guidance network. Unlike the work in \cite{romero2014fitnets, hinton2015distilling, parisotto2016actor, zagoruyko2016paying, ba2014deep}, the guidance network shares the same architecture as the target network but is pretrained with full-precision weights and activations.

However, a pre-trained model may not be necessarily optimal or may not be suitable for quantization. As a result, directly using a fixed pretrained model to guide the target network may not produce the best guidance signals. To mitigate this problem, we do not fix the parameters of a pretrained full precision network as in the previous work \cite{zhang2017deep}.

By using the guidance training strategy, we assume that there exist some full-precision models with good generalization performance, and an accurate low-precision model can be obtained by directly performing the quantization on those full-precision models. In this sense, the feature maps of the learned low-precision model should be close to that obtained by directly doing quantization on the full-precision model. To achieve this, essentially, in our learning scheme, we can jointly train the full-precision and low-precision models. This allows these two models adapt to each other. We even find by doing so the performance of the full-precision model can be slightly improved in some cases.

Formally, let ${{\bf{W}}_{full}}$ and ${{\bf{W}}_{low}}$ be the full-precision model and low-precision model, respectively. Let $\mu ({\bf{x}};{{\bf{W}}_{{full}}})$ and $\nu ({\bf{x}};{{\bf{W}}_{{low}}})$ be the nested feature maps (e.g., activations) of the full-precision model and low-precision model, respectively. To create the guidance signal, we may require that the nested feature maps from the two models should be similar. However,  $\mu ({\bf{x}};{{\bf{W}}_{{full}}})$ and $\nu ({\bf{x}};{{\bf{W}}_{{low}}})$  is usually not directly comparable since one is full precision and the other is low-precision.

To link these two models,  we can directly quantize the weights and activations of the full-precision model by equations (\ref{eq:quan-weigtht}) and (\ref{eq:quan-activations}). For simplicity, we denote the quantized feature maps by  $Q(\mu ({\bf{x}};{{\bf{W}}_{{full}}}))$. Thus, $Q(\mu ({\bf{x}};{{\bf{W}}_{{full}}}))$ and  $\nu ({\bf{x}};{{\bf{W}}_{{low}}})$ will become comparable. Then we can define the guidance loss as:
\begin{equation}
	R({{\bf{W}}_{full}},{{\bf{W}}_{low}}) = \frac{1}{2}\parallel Q(\mu ({\bf{x}};{{\bf{W}}_{{full}}})) - \nu ({\bf{x}};{{\bf{W}}_{{low}}}){\parallel^2},
\end{equation}
where $\parallel\cdot\parallel$ denotes some proper norms.

Let ${L_{{\theta _1}}}$ and ${L_{{\theta _2}}}$ be the cross-entropy classification losses for the full-precision and low-precision model, respectively. The guidance loss will be added to ${L_{{\theta _1}}}$ and ${L_{{\theta _2}}}$, respectively, resulting in two new objectives for the two networks, namely
\begin{equation} \label{eq:objective1}
	L_1({{\bf{W}}_{full}})  = {L_{{\theta _1}}} + \lambda R({{\bf{W}}_{full}},{{\bf{W}}_{low}}).
\end{equation}
and
\begin{equation} \label{eq:objective2}
	L_2({{\bf{W}}_{low}})  = {L_{{\theta _2}}} +  \lambda R({{\bf{W}}_{full}},{{\bf{W}}_{low}}).
\end{equation}
where $\lambda$ is a balancing parameter. Here, the guidance loss $R$ can be considered as some regularization on ${L_{{\theta _1}}}$ and ${L_{{\theta _2}}}$.

In the learning procedure, both ${{\bf{W}}_{full}}$ and ${{\bf{W}}_{low}}$ will be updated by minimizing $L_1({{\bf{W}}_{full}})$ and $L_2({{\bf{W}}_{low}})$ separately, using a mini-batch stochastic gradient descent method. The detailed algorithm is shown in Algorithm \ref{algo:one-mutual learning}. A high-bit precision model $M_{low}^K$ is used as an initialization of $M_{low}^k$, where $K>k$. Specifically, for the full-precision model, we have $K=32$. Relying on $M_{full}$, the weights and activations of $M_{low}^k$ can be initialized by equations (\ref{eq:quan-weigtht}) and (\ref{eq:quan-activations}), respectively.

Note that the training process of the two networks are different.
When updating ${{\bf{W}}_{low}}$ by minimizing $L_2({{\bf{W}}_{low}})$, we use full-precision model as the initialization and apply the forward-backward propagation rule in Section \ref{sec:baseline}  to fine-tune the model. When updating ${{\bf{W}}_{full}}$ by minimizing $L_1({{\bf{W}}_{full}})$, we use conventional forward-backward propagation to fine-tune the model.

\begin{algorithm}[]
	\KwIn{Training data $\{ ({{\bf{x}}_i},y_i)\}_{i=1}^N$; A pre-trained 32-bit full-precision model $M_{full}$; A $k$-bit precision model $M_{low}^k$.}
	\KwOut{A low-precision deep model $M^k_{low}$ with weights and activations being quantized into $k$ bits.}
	Initialize $M_{low}^k$ based on $M_{full}$;\\
	\For{ $\mathrm{epoch} = 1,...,L$}
	{
		\For{ $t = 1,...T$}
		{
			Randomly sample a mini-batch data;\\
			Quantize the weights ${{\bf{W}}_{low}}$  and activations into $k$-bit by minimizing $L_2({{\bf{W}}_{low}})$\;
			Update $M_{full}$ by minimizing $L_1({{\bf{W}}_{full}})$\;
		}

	}
	\caption{Guided training with a full-precision network for $k$-bit quantization}
	\label{algo:one-mutual learning}
\end{algorithm}

\subsection{Remark on the proposed methods}
The proposed three approaches tackle the difficulty in training a low-precision model with different strategies. They can be applied independently. However, it is also possible to combine them together. For example, we can apply the progressive quantization to any of the steps in the two-stage approach; we can also apply the guided training to any sub-step in the progressive training. Detailed analysis on possible combinations will be experimentally evaluated in the experiment section.

\subsection{Implementation details} \label{sec:implementation}

In all the three methods, we quantize the weights and activations of all layers except that the input data are kept to 8-bit. Furthermore, to promote convergence, we propose to add a scalar layer after the last fully-connected layer before feeding the low-bit activations into the softmax function for classification. The scalar layer has only one trainable small scalar parameter and is initialized to 0.01 in our approach.

During training, we randomly crop 224x224 patches from an image or its horizontal flip, with the per-pixel mean subtracted. We don't use any further data augmentation in our implementation. We adopt batch normalization (BN)~\cite{ioffe2015batch} after each convolution before activation. For pretraining the full-precision baseline model, we use Nesterov SGD and batch size is set to 256. The learning rate starts from 0.01 and is divided by 10 every 30 epochs. We use a weight decay 0.0001 and a momentum 0.9. For weights and activations quantization, the initial learning rate is set to 0.001 and is divided by 10 every 10 epochs. We use a simple single-crop testing for standard evaluation. Following~\cite{zagoruyko2016paying}, for ResNet-50, we add only two guidance losses in the 2 last groups of residual blocks. And for AlexNet, we add two guidance losses in the last two fully-connected layers.

	 \section{Experiment}

To investigate the performance of the proposed methods, we conduct experiments on Cifar100 and ImageNet datasets. Two representative networks, different precisions AlexNet and ResNet-50 are evaluated with top-1 and top-5 accuracy reported. We use a variant of AlexNet structure~\cite{krizhevsky2012imagenet} by removing dropout layers and add batch normalization after each convolutional layer and fully-connected layer. This structure is widely used in previous works~\cite{zhou2016dorefa, zhu2016trained}.  We analyze the effect of the guided training approach, two-stage optimization and the progressive quantization in details in the ablation study. Seven methods are implemented and compared:

\begin{enumerate}
\item ``\textbf{Baseline}'': We implement the baseline model based on DoReFa-Net as described in Section~\ref{sec:baseline}.

\item ``\textbf{TS}'': We apply the two-stage optimization strategy described in  Sec.~\ref{sec:two-stage} and Algorithm~\ref{algo:two-stage} to quantize the weights and activations. We denote the first stage as \textbf{Stage1} and the second stage as \textbf{Stage2}.

\item ``\textbf{PQ}'': We apply the progressive quantization strategy described in  Sec.~\ref{sec:progressive} and Algorithm~\ref{algo:progressive optimization} to continuously quantize weights and activations simultaneously from high-precision (\ie, 32-bit) to low-precision.

\item ``\textbf{Guided}'': We implement the guided training approach as described in Sec.~\ref{sec:mutual} and Algorithm~\ref{algo:one-mutual learning} to independently investigate its effect on the final performance.

\item ``\textbf{PQ+TS}'': We further combine \textbf{PQ} and \textbf{TS} together to see whether their combination can improve the performance.

\item ``\textbf{PQ+TS+Guided}'': This implements the full model by combining \textbf{PQ}, \textbf{TS} and \textbf{Guided} modules together.

\item ``\textbf{PQ+TS+Guided**}'': Based on PQ\-+\-TS\-+\-Guided, we use full-precision weights for the first convolutional layer and the last fully-connected layer following the setting of~\cite{zhu2016trained, zhou2016dorefa} to investigate its sensitivity to the proposed method.
\end{enumerate}

\subsection{Evaluation on ImageNet}

We further train and evaluate our model on ILSVRC2012~\cite{russakovsky2015imagenet}, which includes over 1.2 million images and 50 thousand validation images. We report 4-bit and 2-bit precision accuracy for both AlexNet and ResNet-50. The sequence of bit-width precisions are set as $\{32, 8, 4, 2\}$. The results of INQ~\cite{zhou2017incremental} are directly cited from the original paper. We did not use the sophisticated image augmentation and more details can be found in Sec.~\ref{sec:implementation}. We compare our model to the 32-bit full-precision model, INQ, DoReFa-Net and the baseline approach described in Sec.~\ref{sec:baseline}. For INQ, only the weights are quantized. For DoReFa-Net, the first convolutional layer uses the full-precision weights and the last fully-connected layer use both full-precision weights and activations.

\emph{\textbf{Results on AlexNet:}}
The results for AlexNet are listed in Table~\ref{tab:AlexNet}. Compared to competing approaches, we achieve steadily improvement for 4-bit and 2-bit settings. This can be attributed to the effective progressive optimization and the knowledge from the full-precision model for assisting the optimization process. Furthermore, our 4-bit full model even outperforms the full-precision reference by 0.7\% on top-1 accuracy. This may be due to the fact that on this data, we may not need a model as complex as the full-precision one. However, when the expected bit-width decrease to 2-bit, we observe obvious performance drop compared to the 32-bit model while our low-bit model still brings 2.8\% top-1 accuracy increase compared to the \emph{Baseline} method.

\emph{\textbf{Results on ResNet-50:}}
The results for ResNet-50 are listed in Table~\ref{tab:ResNet-50}. For the full-precision model, we implement it using Pytorch following the re-implementation provided by Facebook\footnote{\url{https://github.com/facebook/fb.resnet.torch}}.
Comparatively, we find that the performance are approximately consistent with the results of AlexNet. Similarly, we observe that our 4-bit full model is comparable with the full-precision reference with no loss of accuracy. When decreasing the precision to 2-bit, we achieve promising improvement over the competing \emph{Baseline} even though there's still an accuracy gap between the full-precision model. Similar to the AlexNet on ImageNet dataset, we find our 2-bit full model improves more comparing with the 4-bit case. This phenomenon shows that when the model becomes more difficult to optimize, the proposed approach turns out to be more effective in dealing with the optimization difficulty.
To better understand our model, we also draw the process of training for 2-bit ResNet-50 in Figure~\ref{fig:resnet_2bits} and more analysis can be referred in Sec.~\ref{sec:ablation}.

\begin{table*}[!tbp]
		\centering
		\scalebox{0.7}
		{
			\begin{tabular}{c c c | c c c | c c c }
				\hline
                 Accuracy & Full precision &5-bit (INQ) &4-bit (DoReFa-Net)  & 4-bit (Baseline)  & 4-bit (PQ+TS+Guided) &2-bit (DoReFa-Net) &2-bit (Baseline) & 2-bit (PQ+TS+Guided)\\\hline
                 Top1 &57.2\%  &57.4\%  &56.2\%  &56.8\%  &\bf{58.0}\%  &48.3\%  &48.8\%  &\bf{51.6}\% \\
                 Top5 &80.3\%  &80.6\%  &79.4\% &80.0\% &\bf{81.1}\% &71.6\%  &72.2\%  &\bf{76.2}\%  \\\hline

			\end{tabular}}
			\caption{Top1 and Top5 validation accuracy of AlexNet on ImageNet.}
			\label{tab:AlexNet}
\end{table*}

\begin{table*}[!tbp]
	\centering
	\scalebox{0.7}
	{
		\begin{tabular}{c c c| c c c | c c c }
			\hline
			Accuracy & Full precision &5-bit (INQ) &4-bit (DoReFa-Net) & 4-bit (Baseline)  & 4-bit (PQ+TS+Guided) &2-bit (DoReFa-Net) &2-bit (Baseline) & 2-bit (PQ+TS+Guided)\\\hline
			Top1 &75.6\%  &74.8\%  &74.5\%  &75.1\%  &\bf{75.7}\%  &67.3\%  &67.7\%  &\bf{70.0}\%  \\
			Top5 &92.2\%  &91.7\%  &91.5\%  &91.9\%  &\bf{92.0}\% &84.3\% &84.7\%  &\bf{87.5}\%  \\\hline

		\end{tabular}}
		\caption{Top1 and Top5 validation accuracy of ResNet-50 on ImageNet.}
		\label{tab:ResNet-50}
	\end{table*}

\subsection{Evaluation on Cifar100}
Cifar100 is an image classification benchmark containing images of size 32x32 in a training set of 50,000 and
a test set of 10,000. We use the AlexNet for our experiment. The quantitative results are reported in  Table~\ref{tab:cifar100_AlexNet}. From the table, we can observe that the proposed approach steadily outperforms the competing method DoReFa-Net. Interestingly, the accuracy of our 4-bit full model also surpasses its full precision model. We speculate that this is due to 4-bit weights and activations providing the right model capacity and preventing overfitting for the networks.

\begin{table*}[!tbp]
	\centering
	\scalebox{0.76}
	{
		\begin{tabular}{c c |c  c c | c c c}
			\hline
			Accuracy & Full precision &4-bit (DoReFa-Net)  & 4-bit (Baseline) & 4-bit (PQ+TS+Guided) &2-bit (DoReFa-Net) &2-bit (Baseline) &2-bit (PQ+TS+Guided)\\\hline
			Top1 &65.4\%  &64.9\% &65.0\%   &\bf{65.8}\%  &63.4\% &63.9\%  &\bf{64.6}\%  \\
			Top5 &88.3\% &88.5\%  &88.5\%   &\bf{88.6}\% &87.5\%  &87.6\%  &\bf{87.8}\% \\\hline

		\end{tabular}}
		\caption{Top1 and Top5 validation accuracy of AlexNet on Cifar100.}
		\label{tab:cifar100_AlexNet}
	\end{table*}

\subsection{Ablation study} \label{sec:ablation}
In this section, we analyze the effects of different components of the proposed model.

\begin{figure}
	\centering
	\resizebox{0.9\linewidth}{!}
	{
		\begin{tabular}{c}
			\includegraphics{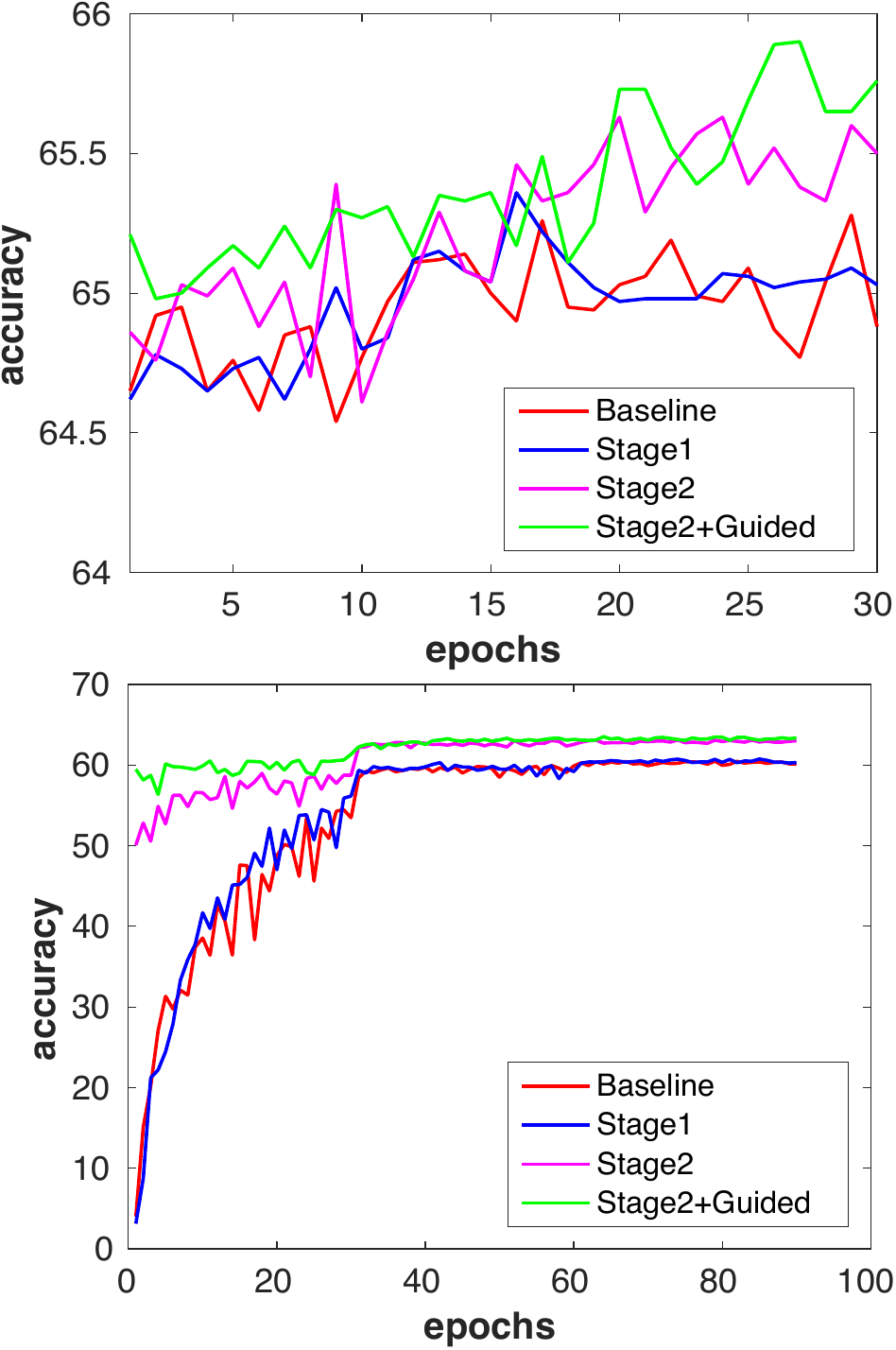}
		\end{tabular}
	}
	\caption{Validation accuracy of 4-bit AlexNet on Cifar100 using (a): the fine-tuning strategy; (b): learning from scratch strategy. \emph{Stage2+Guided} means we combine the methods \emph{Stage2} and \emph{Guided} together during optimization to investigate the effect of the guided training on the final performance.}
	\label{fig:cifar}
\end{figure}

\vspace{1mm}
\noindent\textbf{\emph{Learning from scratch vs. Fine-tuning:}}
To analyze the effect, we perform comparative experiments on Cifar100 with AlexNet using learning from scratch and fine-tuning strategies. The results are shown in Figure~\ref{fig:cifar}, respectively. For convenience of exposition, this comparison study is performed based on method \emph{TS}.
First, we observe that the overall accuracy of fine-tuning from full-precision model is higher than that of learning from scratch. This indicates that the initial point for training low-bitwidth model is crutial for obtaining good accuracy.
In addition, the gap between the \emph{Baseline} and \emph{TS} is obvious (\ie, 2.7 \% in our experiment) with learning from scratch. This justifies that the two-stage optimization strategy can effectively help the model converge to a better local minimum.

\begin{table}[!tbp]
	\centering
	\scalebox{1.0}
	{
		\begin{tabular}{c c c}
			\hline
			Method &top-1  &top-5 \\\hline
			 4-bit (TS) &57.7\% &81.0\%\\
			 4-bit (PQ)  &57.5\%  &80.8\%\\
			 4-bit (PQ+TS) &57.8\% &80.8\% \\
			 4-bit (Guided) &57.3\% &80.4\%\\
			 4-bit (PQ+TS+Guided)  &58.0\% &81.1\%\\
			 4-bit (PQ+TS+Guided**)  &\bf{58.1}\% &\bf{81.2}\% \\\hline
			 2-bit (TS)  &50.7\% &74.9\% \\
			 2-bit (PQ)  &50.3\% &74.8\% \\
			 2-bit (PQ+TS)  &50.9\% &74.9\% \\
			 2-bit (Guided) &50.0\%  &74.1\% \\
			 2-bit (PQ+TS+Guided)  &51.6\% &76.2\% \\
			 2-bit (PQ+TS+Guided**) &\bf{52.5}\% &\bf{77.3}\% \\\hline

		\end{tabular}}
		\caption{Evaluation of different components of the proposed method on the validation accuracy with AlexNet on ImageNet.}
		\label{tab:AlexNet_ablation}
	\end{table}

\begin{table}[!tbp]
	\centering
	\scalebox{1.0}
	{
		\begin{tabular}{c c c}
			\hline
			Method &top-1  &top-5 \\\hline
			4-bit (TS) &75.3\% &91.9\%\\
			4-bit (PQ)  &75.4\%  &91.8\%\\
			4-bit (PQ+TS) &75.5\% &92.0\% \\
			4-bit (Guided) &75.3 \% &91.7\% \\
			4-bit (PQ+TS+Guided)  &75.7\% &92.0\%\\
			4-bit (PQ+TS+Guided**)  &\bf{75.9}\% &\bf{92.4}\% \\\hline
			2-bit (TS)  &69.2\% &87.0\% \\
			2-bit (PQ)  &68.8\% &86.9\% \\
			2-bit (PQ+TS)  &69.4\% &87.0\% \\
		    2-bit (Guided) &69.0\%  & 86.8\%\\
			2-bit (PQ+TS+Guided)  &70.0\% &87.5\% \\
			2-bit (PQ+TS+Guided**) &\bf{70.8}\% &\bf{88.3}\% \\\hline

		\end{tabular}}
			\caption{Evaluation of different components of the proposed method on the validation accuracy with ResNet-50 on ImageNet.}
			\label{tab:ResNet-50_ablation}
\end{table}

\vspace{1mm}
\noindent\textbf{{\emph{The effect of quantizing all layers:}}}
This set of experiments is performed to analyze the influence for quantizing the first convolutional layer and the last fully-connected layer. Several previous works~\cite{zhu2016trained} argue to keep these two layers precision as 32-bit floating points to decrease accuracy loss. By comparing the results of \emph{PQ+TS+Guided**} and \emph{PQ+TS+Guided} in Table~\ref{tab:AlexNet_ablation} and Table~\ref{tab:ResNet-50_ablation}, we notice that the accuracy gap between the two settings is not large, which indicates that our model is not sensitive to the precision of these two layers. It can be attributed to two facts. On one hand, fine-tuning from 32-bit precision can drastically decrease the difficulty for optimization. On the other hand, the progressive optimization approach as well as the guided training strategy further ease the instability during training.

\begin{figure}[!htb]
	\centering
	\resizebox{0.9\linewidth}{!}
	{
		\begin{tabular}{c}
			\includegraphics{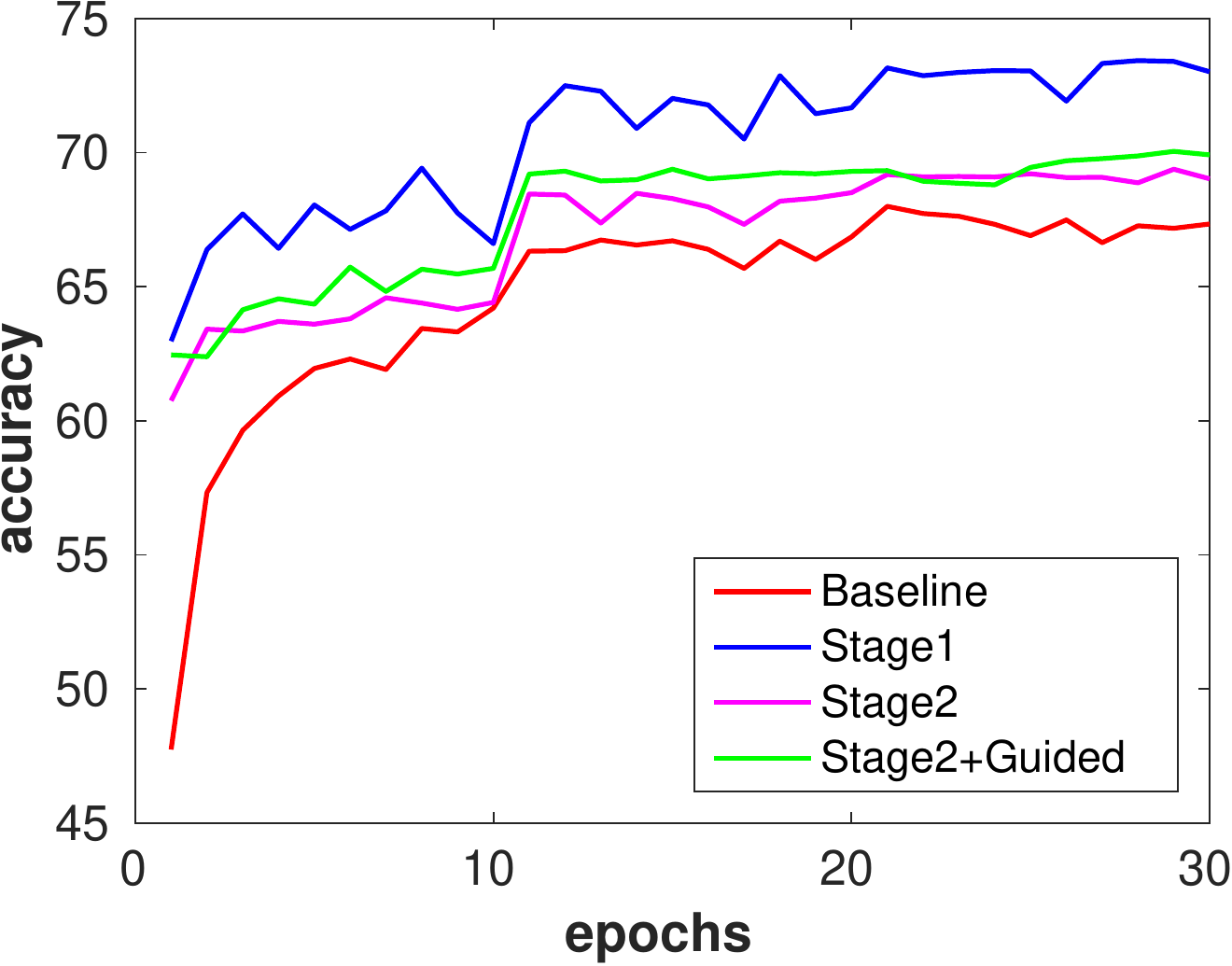}
		\end{tabular}
	}
	\caption{Validation accuracy of 2-bit ResNet-50 on ImageNet. \emph{Stage2+Guided} means we combine the methods \emph{Stage2} and \emph{Guided} together during training.}
	\label{fig:resnet_2bits}
\end{figure}

\vspace{1mm}
\noindent\textbf{{\emph{The effect of the two-stage optimization strategy:}}}
We further analyze the effect of each stage in the \emph{TS} approach in Figure~\ref{fig:cifar} and Figure~\ref{fig:resnet_2bits}.
We take the 2-bitwidth ResNet-50 on ImageNet as an example. In Figure~\ref{fig:resnet_2bits}, \emph{Stage1} has the minimal loss of accuracy. As for the \emph{Stage2}, although it incurs apparent accuracy decrease in comparison with that of the \emph{Stage1}, its accuracy is consistently better than the results of \emph{Baseline} in every epoch. This illustrates that progressively seeking for the local minimum point is crutial for final better convergence. We also conduct additional experiments on Cifar100 with 4-bit AlexNet. Interestingly, taking the model of \emph{Stage1} as the initial point, the results of \emph{Stage2} even have relative increase using two different training strategies as mentioned above. This can be interpreted by that further quantizing the activations impose more regularization on the model to overcome overfitting.
Overall, the two-step optimization strategy still performs steadily better than the Baseline method which proves the effectiveness of this simple mechanism.

\begin{figure}[!htb]
	\centering
	\resizebox{0.9\linewidth}{!}
	{
		\begin{tabular}{c}
			\includegraphics{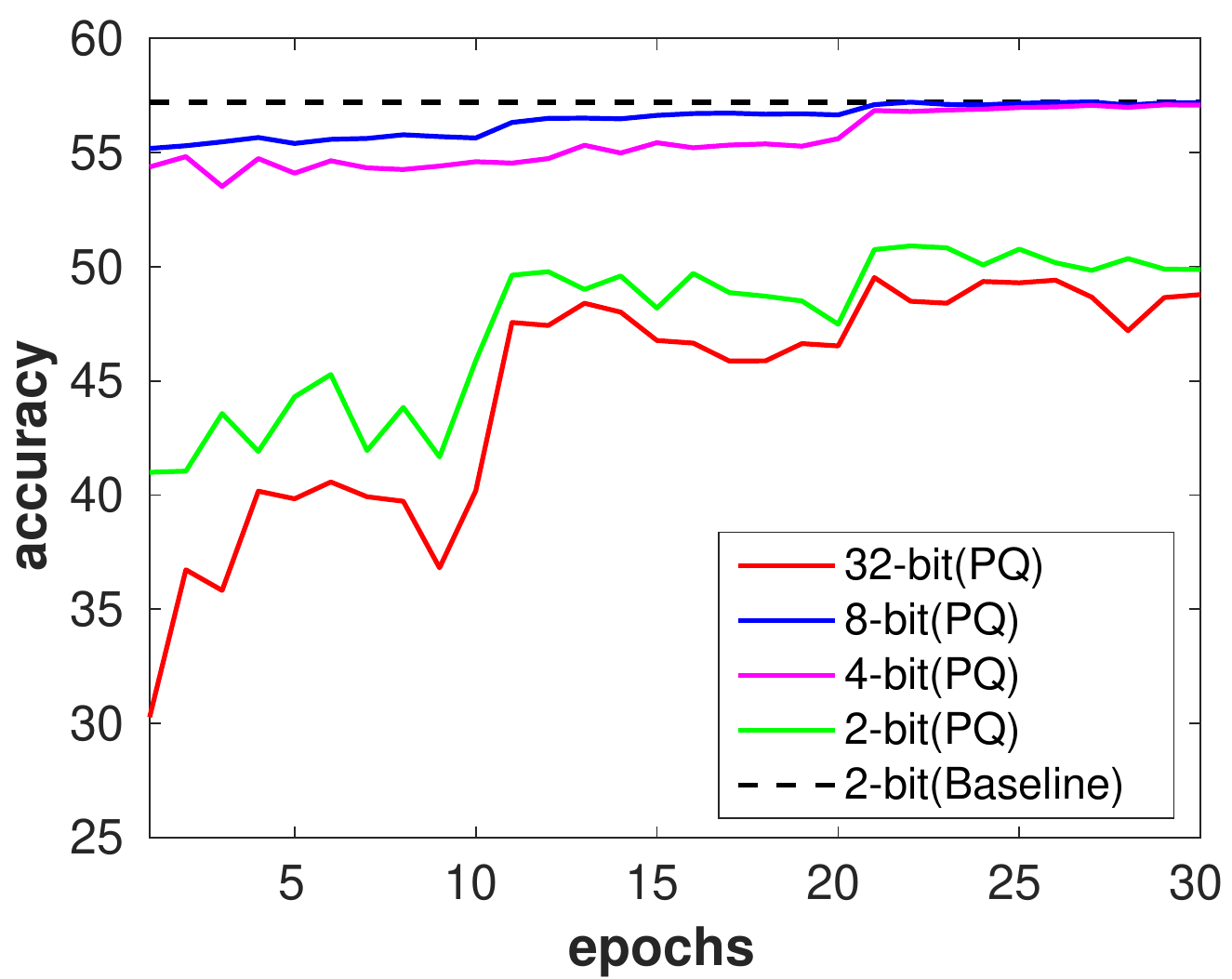}
		\end{tabular}
	}
	\caption{Validation accuracy of the progressive quantization approach using AlexNet on ImageNet.}
	\label{fig:progressive}
\end{figure}

\vspace{1mm}
\noindent\textbf{{\emph{The effect of the progressive quantization strategy:}}} What's more, we also separately explore the progressive quantization (\ie, \emph{PQ}) effect on the final performance.
In this experiment, we apply AlexNet on the ImageNet dataset.
We continuously quantize both weights and activations simultaneously from 32-bit$\to$8-bit$\to$4-bit$\to$2-bit and explictly illustrate the accuracy change process for each precision in Figure~\ref{fig:progressive}. The quantitative results are also reported in Table~\ref{tab:AlexNet_ablation}  and Table~\ref{tab:ResNet-50_ablation}. From the figure we can find that for the 8-bit and 4-bit, the low-bit model has no accuracy loss with respect to the full precision model. However, when quantizing from 4-bit to 2-bit, we can observe significant accuracy drop.
Despite this, we still observe $1.5\%$ relative improvement by comparing the top-1 accuracy over the 2-bit baseline, which proves the effectiveness of the proposed strategy. It is worth noticing that the accuracy curves become more unstable when quantizing to lower bit. This phenomenon is reasonable since the precision becomes lower, the value will change more frequently during training.

\vspace{1mm}
\noindent\textbf{{\emph{The effect of the jointly guided training:}}}
We also investigate the effect of the guided joint training approach explained in Sec.~\ref{sec:mutual}. By comparing the results in Table~\ref{tab:AlexNet_ablation} and Table~\ref{tab:ResNet-50_ablation}, we can find that \emph{Guided} method steadily improves the \emph{baseline} method by a promising margin. This justifies the low-precision model can always benefit by learning from the full-precision model.
What's more, we can find \emph{PQ+TS+Guided} outperforms \emph{PQ+TS} in all settings. This shows that the guided training strategy and the progressive learning mechanism can benefit from each other for further improvement.

\begin{figure}[!htb]
	\centering
	\resizebox{0.9\linewidth}{!}
	{
		\begin{tabular}{c}
			\includegraphics{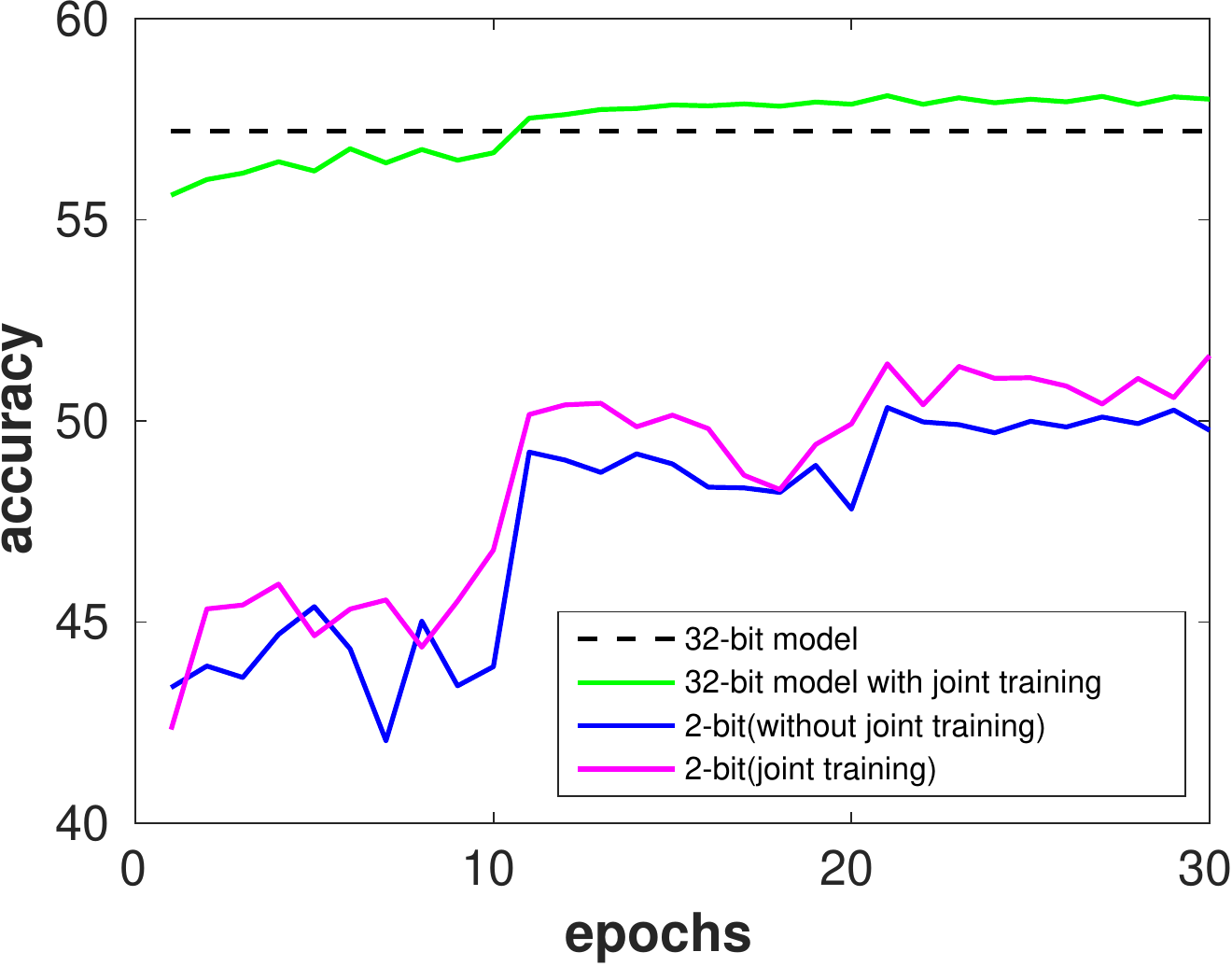}
		\end{tabular}
	}
	\caption{The effect of the joint training strategy using AlexNet on ImageNet.}
	\label{fig:mutual}
\end{figure}

\vspace{1mm}
\noindent\textbf{{\emph{Joint vs. without joint:}}}
We further illustrate the joint optimization effect on guided training in Figure~\ref{fig:mutual}. For explaning convenience, we implement it based on the method \emph{Stage2+Guided} and report the 2-bit AlexNet top-1 validation accuracy on ImageNet. From the figure, we can observe that both the full-precision model and its low-precision counterpart can benefit from learning from each other. In contrast, if we keep the full-precision model unchanged, apparent performance drop is observed. This result strongly supports our assumption that the high-precision and the low-precision models should be jointly optimized in order to obtain the optimal gradient during training. The improvement on the full-precision model may due to the ensemble learning with the low-precision model and similar observation is found in~\cite{zhang2017deep} but with different task.

	 \section{Conclusion}
In this paper, we have proposed three novel approaches to solve the optimization problem for quantizing the network with both low-precision weights and activations. We first propose a two-stage approach to quantize the weights and activations in a two-step manner. We also observe that continuously quantize from high-precision to low-precision is also beneficial to the final performance. To better utilize the knowledge from the full-precision model, we propose to jointly learn the low-precision model and its full-precision counterpart to optimize the gradient problem during training. Using 4-bit weights and activations for all layers, we even outperform the performance of the 32-bit model on ImageNet and Cifar100 with general frameworks. 

\small
\bibliographystyle{ieee}
\bibliography{reference}

\end{document}